\documentclass[runningheads]{llncs}
\usepackage{graphicx}
\graphicspath{{images/}}

\usepackage{times}
\usepackage{latexsym}

\usepackage[bottom,flushmargin,hang,multiple]{footmisc}
\usepackage{dblfloatfix}
\usepackage{tabulary}

\usepackage{microtype}

\usepackage{amsmath}
\usepackage{amssymb}
\usepackage{gensymb}

\usepackage[utf8]{inputenc}
\usepackage[english]{babel}
\usepackage{graphicx}
\usepackage{upgreek}
\usepackage{authblk}
\usepackage{todonotes}
\usepackage{multirow}
\usepackage{url}

\usepackage{tabularx}
\usepackage[normalem]{ulem}
\useunder{\uline}{\ul}{}

\usepackage[utf8]{inputenc}

\usepackage[hang]{footmisc}
\bfseries

\begin{document}

\title{WikiMulti: a Corpus for Cross-Lingual Summarization}

\author{Pavel Tikhonov\inst{1} \and
Valentin Malykh\inst{2}}
\authorrunning{P. Tikhonov \& V. Malykh}
\institute{Independent researcher\\
\email{pav3370@yandex.ru} \and
Kazan Federal University, Kazan, Russia \\
\email{valentin.malykh@phystech.edu}
}

\makeatletter
\renewcommand\@makefntext[1]{\leftskip=2em\hskip-2em\@makefnmark#1}
\makeatother

\makeatletter
\newcommand{\specialcell}[2][c]{%
  \begin{tabular}[#1]{@{}c@{}}#2\end{tabular}}
\makeatother

\makeatletter
\newcommand{\thickhline}{%
    \noalign {\ifnum 0=`}\fi \hrule height 1pt
    \futurelet \reserved@a \@xhline
}
\makeatother

\maketitle
\begin{abstract}
Cross-lingual summarization (CLS) is the task to produce a summary in one particular language for a source document in a different language.
We introduce WikiMulti - a new dataset for cross-lingual summarization based on Wikipedia articles in 15 languages. As a set of baselines for further studies, we evaluate the performance of existing cross-lingual abstractive summarization methods on our dataset. We make our dataset publicly available here: \url{https://github.com/tikhonovpavel/wikimulti}

\end{abstract}

% \section{Introduction}

\section{Introduction}

Automatic summarization is one of the central problems in Natural Language Processing (NLP) posing several challenges relating to understanding (i.e. identifying important content) and generation (i.e. aggregating and rewording the identified content into a summary). Of the many summarization paradigms that have been identified over the years, single-document summarization has consistently garnered attention. Given an input text (typically a long document or article), the goal is to generate a smaller, concise piece of text that conveys the key information of the input text. There are two main approaches to automatic text summarization: extractive and abstractive. Extractive methods chop out one or more segments from the input text and concatenate them to produce a summary. These methods were dominant in the early era of summarization, but they suffer from some limitations, including weak coherence between sentences, inability to simplify complex and long sentences, and unintended repetition.
Abstractive text summarization is the task of generating a short and concise summary that captures the salient ideas of the source text.

Despite the presence of a large number of datasets for abstractive summarization~\cite{sandhaus2008new,nallapati2016abstractive,narayan2018don}, the vast majority of them are focused on \textit{mono-lingual} summarization.

However, there exists a number of summarization datasets including several languages.
The task for summarization on several languages could be stated in two significantly different ways. The one is called \textit{cross-lingual} and other is \textit{multi-lingual}. In the case of multi-lingual datasets, the corpus is collected in several languages, but there is no requirement for an alignment, in the sense that the documents in one language may not correspond to the documents in any other language. The systems trained on such corpora are targeted to produce summary of a document of the same language, e.g. a system should make summaries for Portuguese documents in form of the paragraphs in Portuguese. \textit{Multiling'13 and '15}~\cite{giannakopoulos2013multi,elhadad2013multi,giannakopoulos2015multiling},
MLSUM~\cite{scialom2020mlsum}, and XL-Sum~\cite{hasan2021xl}
are examples of a multi-lingual datasets.

In the case of a cross-lingual dataset, the corpus have to be aligned between the languages. For example, the document in English should have a summary in Portuguese. The systems trained on such datasets should be able to make a summary in another language regarding the language of the input document.

There were a few attempts for addressing the problem of cross-lingual summarization \cite{ladhak2020wikilingua,nguyen2019global}. Among them, only \cite{ladhak2020wikilingua} is the only one of the datasets which is large and addresses the problem of cross-lingual summarization.
However, this dataset contains only short articles for a few topics.

% There were a few attempts for addressing the problem of cross-lingual summarization. The dataset described in \cite{nguyen2019global}
% % ~\cite{giannakopoulos2013multi,elhadad2013multi,li2013multi,nguyen2019global,giannakopoulos2015multiling} 
% is of small size, while the only one of the datasets which is large and address the problem of cross-lingual summarization is WikiLingua~\cite{ladhak2020wikilingua}. %In this dataset the authors crawled WikiHow site to collect a corpus, using annotations for images.

This further opens up avenues to explore new approaches for cross-lingual summarization, which are currently understudied. We present a novel dataset \textbf{WikiMulti} consisting of Wikipedia articles and summaries in 15 languages.
With the dataset in hand, we evaluate several approaches for cross-lingual summarization to establish the baselines.

This paper is structured as follows: in Sec.~\ref{sec:related} we review existing datasets on multi- and cross-lingual summarization; in Sec.~\ref{sec:dataset} we describe WikiMulti, the presented dataset; Sec.~\ref{sec:experiments} is devoted to the description of the baselines for this dataset; Sec.~\ref{sec:results} contains the results for the baselines, while Sec.~\ref{sec:conclusion} concludes the paper.
% Most of existing multilingual datasets mostly consists of parallel pairs which is good for translation, but not for summarization.

% Однако существует проблема кросс-языковой суммаризации. Это важно по тому то. Эта проблема возникает в таких-то случаях\\
% И датасетов адресующих эту проблему практически нет\\
% Поэтому на сцену выходим мы и представляем наш классный датасет\\

\section{Existing Datasets}
\label{sec:related}
In this section, we take a closer look at the multi- and cross-lingual summarization datasets. The statistics on these datasets provided in Tab.~\ref{tab:comparison}.

\begin{table*}[tbh!]
\centering
\begin{tabular}{l|c|c|c|c}

 \thickhline
                       & Num languages & Avg num summaries & Avg summary length & Avg article length \\ \hline
\textsc{MultiLing'15}  & 40      & 30        & 185       & 4,111     \\ \hline
% \textsc{XL-Sum}      &   44    & 22,847  &      &            \\ \hline
\textsc{MLSUM}      & 5       & 314,208  &   34   &    812        \\ \hline
\thickhline
\textsc{Global Voices} & 15      & 1,456     & 51        & 359       \\ \hline
\textsc{WikiLingua}    & 18      & 42,783    & 39        & 391       \\ \hline
\textsc{WikiMulti}     & 15      & 10,467    & 112       & 1078      \\ 

 \thickhline
% \hline
\end{tabular}
\caption{Statistics for existing multi-lingual (top) and cross-lingual (bottom) datasets.}
\label{tab:comparison}
\end{table*}

\subsection{Multi-lingual datasets}
\textbf{Multiling'13}~~\cite{giannakopoulos2013multi,elhadad2013multi} and \textbf{Multiling'15}~\cite{giannakopoulos2015multiling} have been collected at MultiLing Workshops by organizers. The MultiLing’13 dataset includes summaries of 30 Wikipedia articles per language, describing a given topic. For MultiLing’15, an additional 30 documents were collected for evaluation purposes.

\textbf{MLSUM}~\cite{scialom2020mlsum}: A dataset obtained from online newspapers. It contains 1.5 million article/summary pairs in five different languages, namely, French, German, Spanish, Russian, and Turkish.

% The Cross-Lingual NLI Corpus
\textbf{XL-Sum}~\cite{hasan2021xl}:
 A dataset containing 1 million article-summary pairs in 44 languages, being the first publicly available abstractive summarization dataset for many of them.
 The dataset covers 44 languages ranging from low to high-resource.

\subsection{Cross-lingual datasets}

\textbf{Global Voices}~\cite{nguyen2019global}: authors collected descriptions of news articles provided by Global Voices site creators (it's an international, multilingual community of writers, translators, academics, and digital rights activists.). This dataset supports 15 languages, however, 10 of them have less than 1,000 articles.

\textbf{WikiLingua}~\cite{ladhak2020wikilingua}: authors crawled WikiHow site (is an online resource of how-to guides where each page includes multiple methods for completing a multi-step procedural task along with a one-sentence summary of each step).

\section{WikiMulti Dataset}
\label{sec:dataset}
The well-known community collected encyclopedic resource of Wikipedia is a source for many datasets~\cite{chen2017reading,wang2021wikigraphs,srinivasan2021wit} to name a few, due to on the one hand the massive contents with a variety of topics and languages, curation of the content (for the most popular languages), and on the other hand, the permissive Creative Common license\footnote{``Text is available under the Creative Commons Attribution-ShareAlike License 3.0'', \url{https://en.wikipedia.org/wiki/Main_Page}} used throughout the whole Wikipedia.

\begin{figure*}[t]
%   \vspace{70pt}%
    \centering
    \includegraphics[width=\textwidth]{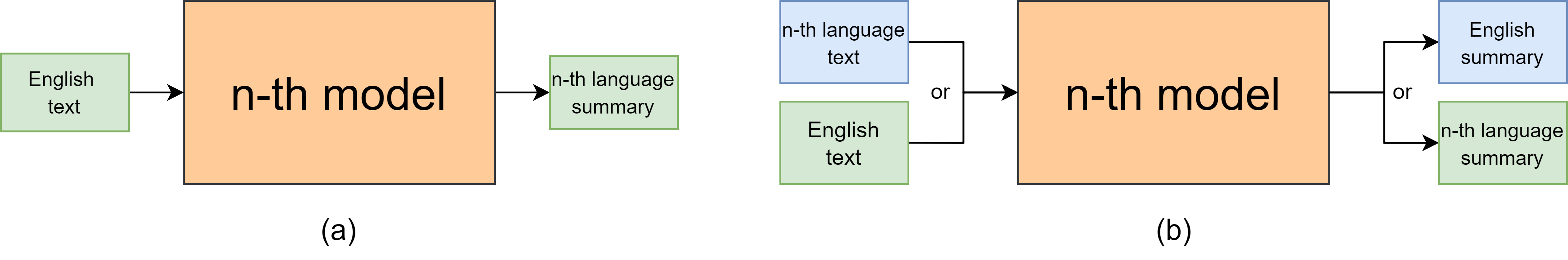}
    \caption{(a) - one-directional approach. (b) - two-directional approach}
        %   For training, a prefix was written, supposedly summarized from such and such into such and such a language
    \label{fig:variants}
\end{figure*}
% \subsection{Mining approach}
Wikipedia project has a concept of so-called Good Article, i.e. the article which is approved by the community as the one describing a specific topic in full detail and well written. One point of this article structure includes the summary as the first paragraph of an article. We decided to build our dataset on this basis.  To produce the corpus, we take a list of Wikipedia's Good  Articles\footnote{\url{https://en.wikipedia.org/wiki/Wikipedia:Good\_articles/all}\\
\url{https://en.wikipedia.org/wiki/Wikipedia:Good\_articles/Social\_sciences\_and\_society}\\
\url{https://en.wikipedia.org/wiki/Wikipedia:Good\_articles/Sports\_and\_recreation}\\ 
\url{https://en.wikipedia.org/wiki/Wikipedia:Good\_articles/Video\_games}\\
\url{https://en.wikipedia.org/wiki/Wikipedia:Good\_articles/Warfare}
} and get a corresponding article in 14 other languages for each article in the list.

Each article belongs to some categories and subcategories. For example category "Language and literature" divided into "Ancient texts", "Comics", "Novels", "Characters and fictional items", etc.
The dataset contains categories from "Architecture – Bridges and tunnels" to "Video game history and development".

A typical Wikipedia article is structured as follows: the first paragraph consisting of 3-7 sentences describes the subject of the article briefly. While the rest of the article contains the details.
We use the first paragraph as a summary and the rest of the article as a text to summarize. The samples from the collected dataset can be seen in Tab.~\ref{tab:erroranalysis}.

\begin{table*}[t!]
  \center{ \footnotesize
    \begin{tabular}{l|p{14cm}}
    %   \thickhline 
      
      % 4306
    %   dsafdsf & f \\
      \thickhline
      L. & Summary\\
      \thickhline
\textsc{en} &     
Outer space, commonly shortened to space, is the expanse that exists beyond Earth and its atmosphere and between celestial bodies. Outer space is not completely empty - it is a hard vacuum containing a low density of particles, predominantly a plasma of hydrogen and helium, as well as electromagnetic radiation, magnetic fields, neutrinos, dust, and cosmic rays. The baseline temperature of outer space, as set by the background radiation from the Big Bang, is 2.7 kelvins ( $-270.45 \degree C$; $-454.81 \degree F$). The plasma between galaxies is thought to account for about half of the baryonic (ordinary) matter in the universe, having a number density of less than one hydrogen atom per cubic metre and a temperature of millions of kelvins. Local concentrations of matter have condensed into stars and galaxies. Studies indicate that 90\% of the mass in most galaxies is in an unknown form, called dark matter, which interacts with other matter through gravitational but not electromagnetic forces. Observations suggest that the majority of the mass-energy in the observable universe is dark energy, a type of vacuum energy that is poorly understood. Intergalactic space takes up most of the volume of the universe, but even galaxies and star systems consist almost entirely of empty space. \\
\hline
\textsc{fr} &     
L'espace désigne les zones de l'Univers situées au-delà des atmosphères et des corps célestes. Il s'agit de l'étendue de densité quasi nulle qui sépare les astres. On parle aussi de vide spatial. Selon les endroits de l'espace désignés, on le qualifie quelquefois d'espace cislunaire, interplanétaire, interstellaire (ou intersidéral) et intergalactique pour désigner plus précisément le vide spatial qui est délimité respectivement par le système Terre-Lune, les planètes, les étoiles et les galaxies. L'espace peut aussi se définir en opposition à l'atmosphère terrestre. \\
\hline
            
\textsc{de} &     
Der Weltraum bezeichnet den Raum zwischen Himmelskörpern. Die Atmosphären von festen und gasförmigen Himmelskörpern (wie Sternen und Planeten) haben keine feste Grenze nach oben, sondern werden mit zunehmendem Abstand zum Himmelskörper allmählich immer dünner. Ab einer bestimmten Höhe spricht man vom Beginn des Weltraums. Im Weltraum herrscht ein Hochvakuum mit niedriger Teilchendichte. Er ist aber kein leerer Raum, sondern enthält Gase, kosmischen Staub und Elementarteilchen (Neutrinos, kosmische Strahlung, Partikel), außerdem elektrische und magnetische Felder, Gravitationsfelder und elektromagnetische Wellen (Photonen). Das fast vollständige Vakuum im Weltraum macht ihn außerordentlich durchsichtig und erlaubt die Beobachtung extrem entfernter Objekte, etwa anderer Galaxien. Jedoch können Nebel aus interstellarer Materie die Sicht auf dahinterliegende Objekte auch stark behindern.  \\
\hline
\textsc{nl} & De ruimte of kosmische ruimte is in de astronomie en voor het onderscheid tussen luchtvaart en ruimtevaart het deel van het heelal op meer dan 100 kilometer van de Aarde. Deze grens is de Kármánlijn, hoewel het geen lijn is, maar een boloppervlak. Er is nog wel discussie of een hoogte van 80 kilometer niet meer voldoet aan relevante natuurkundige criteria.Deep space is het deel van het heelal op grotere afstand dan de Aarde-Maan-lagrangepunten, ruim verder dan de Maan.
De ruimte is geen echt vacuüm, maar bestaat hoofdzakelijk uit plasma van waterstof en helium, elektromagnetische straling (in het bijzonder kosmische achtergrondstraling) en neutrino's. De ruimte bevat zeer weinig atomen van andere elementen (metalen) en stofdeeltjes. De intergalactische ruimte bevat slechts enkele waterstofatomen per kubieke centimeter (in ingeademde lucht zitten ongeveer 1019 atomen per kubieke centimeter). Volgens de meeste theorieën is de ruimte daarnaast rijk aan donkere energie en donkere materie. Ook kunnen er objecten doorheen bewegen, zoals meteoroïden en kometen. \\

\thickhline
    \end{tabular}     
  }
%   \label{tab:samples}
  \caption{Example summaries from WikiMulti for Wikipedia artile ``Outer Space''.}
  \label{tab:erroranalysis}
\end{table*}

% % Main text of the article is divided by sections. \\
% The summary is the first paragraph of an article, the text to summarize is the rest of the article. 

% With the dataset in hand, we evaluate existing approaches for cross-lingual summarization as baselines. We then propose a method for direct crosslingual abstractive summarization, leveraging synthetic data and machine translation as a pre-training step. We show that our method outperforms existing baselines, without relying on translation at inference time.

Our final dataset consists of 22,061 unique English articles. Other languages have, on average, 9,639 articles that align with an article in English. From the Wikipedia list, those with more than 1 million articles were selected. From the list of Wikipedias\footnote{\url{https://en.wikipedia.org/wiki/List_of_Wikipedias}}, those with more than 1 million articles were selected. Several of such Wikipedias were skipped (namely, Waray, Cebuano, Egyptian Arabic) due to the most articles in them have one or two paragraphs.

% Despite the fact that the dataset was collected only for English Good Articles, it can be used for multilingual learning, and not just bilingual, since a large number of articles have more than 1 analogue in another language.

%a large number of articles have more than 1 analogue in another language (to be called not bilingual, but multilingual)

More detailed statistics for our dataset is shown at Tab.~\ref{tab:dataset_statistics} while its comparison to other existing cross-lingual datasets is shown in Tab.~\ref{tab:comparison}.

% \begin{table}
%   \begin{center}{\fontsize{8.5}{11}\selectfont % \footnotesize
%   \begin{tabular}{ l | c c c c } 
%     \thickhline
%     \multirow{2}{*}{Models} & \multicolumn{4}{c}{\% of novel n-grams in generated summaries}  \\
%     & unigrams & bigrams & trigrams & 4-grams  \\ \thickhline 
%     \textsc{lead} & 0.00 & 0.00 & 0.00 & 0.00 \\
%     \textsc{ext-oracle} & 0.00 & 0.00 & 0.00 & 0.00 \\
%     % \textsc{Seq2Seq} & 36.66 & 82.17 & 95.58 & 98.63  \\
%     \textsc{PtGen} & 27.40 & 73.33 & 90.43 & 96.04 \\ 
%     % \textsc{PtGen-Covg} & 25.71 & 70.76 & 88.87 & 95.24 \\ \hline
%     \textsc{ConvS2S} & \textbf{31.26} & \textbf{79.50} & \textbf{94.28} & \textbf{98.10}\\
%     \textsc{T-ConvS2S}  & 30.73 & 79.18 & 94.10 & 98.03 \\ \hline
%     \textsc{gold} & 35.76 & 83.45  & 95.50  & 98.49 \\ \thickhline
%   \end{tabular}}
%   \end{center}
%   \caption{Proportion of novel $n$-grams in summaries generated by various models on the XSum test set.
%     \label{tab:novelngram-xsum-models}}
% \end{table}

\begin{table}[tbh!]
\centering
\begin{tabular}{c|c|c}
\thickhline
% \hline
Language     & Language code    &   Articles \\\hline
English      &    \textsc{en}   &   22061 \\
French       &    \textsc{fr}   &   14625 \\
Spanish      &    \textsc{es}   &   13068 \\
Italian      &    \textsc{it}   &   11847 \\
Russian      &    \textsc{ru}   &   11703 \\
German       &    \textsc{de}   &   11228 \\
Portuguese   &    \textsc{pt}   &   10441 \\
Japanese     &    \textsc{ja}   &   8922 \\
Polish       &    \textsc{pl}   &   8875 \\
Chinese      &    \textsc{zh}   &   8711 \\
Swedish      &    \textsc{sv}   &   8007 \\
Dutch        &    \textsc{nl}   &   7681 \\
Arabic       &    \textsc{ar}   &   7476 \\
Ukranian     &    \textsc{uk}   &   7216 \\
Vietnamese   &    \textsc{vi}   &   5153 \\
\hline
Average & & 9639 \\
\thickhline
\end{tabular}
\caption{Number of articles on different languages in WikiMulti.}
\label{tab:dataset_statistics}
\end{table}

\section{Experiments}
\label{sec:experiments}
In all the experiments we used classic ROUGE scores described in~\cite{lin-2004-rouge} for evaluation in our experiments. We use all the most common variances of ROUGE scores, namely, Precision, Recall, and F-measure for ROUGE-1, ROUGE-2, and ROUGE-L.

\subsection{Baselines}
We evaluate the following baseline approaches for cross-lingual summarization on our data:

\textbf{TextRank+Translate:} we have used TextRank \cite{mihalcea2004textrank} tool to automatically get a summary of the text without using complex models, and then translate summary to target language. Following the recommendation from \cite{ladhak2020wikilingua} we used Amazon translating tool\footnote{\url{https://aws.amazon.com/translate/}} to perform translation.

Also we fine-tuned several models to perform cross-lingual summarization task to do direct cross-lingual learning. Fine tuning on different models might give a better idea of which architectures are best suited. We've used the following models:

\textbf{mBART}~\cite{liu2020multilingual} is a multi-lingual language model that has been trained on large, monolingual corpora in 25 languages. The model uses a shared sub-word vocabulary, encoder, and decoder across all 25 languages, and is trained as a denoising auto-encoder during the pre-training step. mBART is trained once for all languages, providing a set of parameters that can be fine-tuned for any of the language pairs in both supervised and unsupervised settings, without any task-specific or language-specific modifications or initialization schemes.

\textbf{M2M100}~\cite{fan2021beyond} is a multilingual encoder-decoder (seq-to-seq) model primarily intended for translation task. It was originally pre-trained on a dataset that covers thousands of language directions with supervised data, created through large-scale mining. One of the main goals stated by the authors is to focus on a non-English-centric approach: the model can translate directly between any pair of 100 languages.

\textbf{mT5}~\cite{xue2020mt5} is a massive model, a multilingual variant of T5 that was pre-trained on a Common Crawl-based dataset covering 101 languages. The model was trained with "Text-to-Text Transfer Transformer" paradigm which means casting every task, including translation, question answering and classification as feeding the model text as input and training it to generate some target text. This allows to use the same model, loss function, hyperparameters, etc. across diverse set of tasks.
%using the same model to perform different tasks, including translation, question answering and classification.

% \begin{figure}[tbh!]
%     \centering
%     \includegraphics[width=\linewidth]{var2.png}
%     \caption{SumHiS: ranking model (right) + hidden structure discovery model (left).}
%     \label{fig:model_general}
% \end{figure}

To train M2M100 and mBART we took a one-directional approach: train 14 different models using English as source language and summarize English text into one of 14 languages. I.e. for a French-English pair, all texts will be in English and the model will summarize them into French.
% In the case of a cross-lingual dataset, the corpus 056
% must be aligned between the languages. I.e. the 057
% document in English should have a summary in 058
% Portuguese. The systems trained on such datasets 059
% should be able to make a summary in another lan- 060
% guage regarding the language of the input docu- 061
% ment. There were a few attempts for addressing 062
% the problem

To train mT5 we took a different two-directional approach: train 14 different models, but use both English and non-English articles as text to summarize and as summaries 50\% of time. In this case for the same French-English pair, half of the texts will be in English, and the model will summarize them in French, and the other half of the texts will be in French, and the model will summarize them into English.

Figure~\ref{fig:variants} illustrates  these two kinds of approaches.

% Для м2м100 и мбарта вот так: с англ на 14 других
% ДЛя мт5 с того языка на этот и с этого на тот
% А ещё если про сам датасет

\subsection{Experiment Parameters}
We fine-tuned mT5, M2M100 and mBART models for 20k steps on a distributed cluster of 7 Nvidia Tesla P100 GPUs. We used AdamW with cosine learning rate schedule with a linear warmup of 500 steps.

\section{Results and Analysis}
\label{sec:results}
Tab.~\ref{tab:evaluation_results} shows ROUGE scores  for the evaluated baselines. %We observe that the lead baseline performs poorly for this task, unlike in the news domain where it's shown to be a strong baseline \cite{BRANDOW1995674}.

M2M100 showed the highest performance on average, especially compared to mBART and mT5. However, all three M2M100, TextRank+Translate, and mBART have problems with Japanese and Chinese languages, where mT5 is better than all the others.

Also, it is interesting that for the Dutch language, all models show on average a larger ROUGE score than in other languages.

% TextRank+Translate and mBART showed very poor performance on Japanese and Chinese languages.

% mBART very purely on Japanese and Chinese languages too. 
% % Using 

% When comparing the performance of two-direction vs one-direction approaches, using two-directional approach showed no significant difference with one-directional approach. On some one-directinal approach showed a better performance than two-directional.

% Some of the languages showed pure performance: Arabian, Chinese, Japanese perhaps of the ...

% In NL, it seems that the quality is better than in other languages for some reason (why??)

\begin{table}[tbh!]
\centering
% \resizebox{0.4\paperheight}{!}{
\begin{tabular}{c|c|ccccccccc}

 \thickhline
\multicolumn{1}{c|}{\multirow{2}{*}{Model}} &
  \multirow{2}{*}{Language} &
  \multicolumn{3}{c|}{ROUGE-1} &
  \multicolumn{3}{c|}{ROUGE-2} &
  \multicolumn{3}{c}{ROUGE-L} \\ %\cline{3-11} 
\multicolumn{1}{c|}{} &
  &
  \multicolumn{1}{c}{F} &
  \multicolumn{1}{c}{P} &
  \multicolumn{1}{c|}{R} &
  \multicolumn{1}{c}{F} &
  \multicolumn{1}{c}{P} &
  \multicolumn{1}{c|}{R} &
  \multicolumn{1}{c}{F} &
  \multicolumn{1}{c}{P} &
  \multicolumn{1}{c}{R} \\ 

 \thickhline
  
\multirow{13}{*}{\specialcell{\textsc{TextRank}+\\\textsc{Translate}}}  
& \textsc{ar} & 0.11 & 0.10 & \multicolumn{1}{c|}{0.15} & 0.02 & 0.01 & \multicolumn{1}{c|}{0.02} & 0.10 & 0.08 & 0.12 \\
& \textsc{de} & 0.14 & 0.11 & \multicolumn{1}{c|}{0.23} & 0.02 & 0.01 & \multicolumn{1}{c|}{0.04} & 0.12 & 0.09 & 0.20 \\
& \textsc{es} & 0.21 & 0.17 & \multicolumn{1}{c|}{0.28} & 0.04 & 0.03 & \multicolumn{1}{c|}{0.06} & 0.17 & 0.14 & 0.24 \\
& \textsc{fr} & 0.20 & 0.16 & \multicolumn{1}{c|}{0.31} & 0.04 & 0.03 & \multicolumn{1}{c|}{0.08} & 0.17 & 0.13 & 0.27 \\
& \textsc{it} & 0.18 & 0.15 & \multicolumn{1}{c|}{0.27} & 0.03 & 0.02 & \multicolumn{1}{c|}{0.05} & 0.16 & 0.13 & 0.23 \\
& \textsc{ja} & 0.01 & 0.01 & \multicolumn{1}{c|}{0.01} & 0.00 & 0.00 & \multicolumn{1}{c|}{0.00} & 0.01 & 0.01 & 0.01 \\
& \textsc{nl} & 0.18 & 0.14 & \multicolumn{1}{c|}{0.31} & 0.04 & 0.03 & \multicolumn{1}{c|}{0.07} & 0.16 & 0.12 & 0.28 \\
& \textsc{pl} & 0.10 & 0.07 & \multicolumn{1}{c|}{0.19} & 0.02 & 0.01 & \multicolumn{1}{c|}{0.04} & 0.09 & 0.07 & 0.17 \\
& \textsc{pt} & 0.18 & 0.15 & \multicolumn{1}{c|}{0.25} & 0.02 & 0.02 & \multicolumn{1}{c|}{0.04} & 0.15 & 0.13 & 0.22 \\
& \textsc{ru} & 0.08 & 0.06 & \multicolumn{1}{c|}{0.14} & 0.01 & 0.01 & \multicolumn{1}{c|}{0.02} & 0.07 & 0.06 & 0.13 \\
& \textsc{sv} & 0.15 & 0.11 & \multicolumn{1}{c|}{0.26} & 0.02 & 0.01 & \multicolumn{1}{c|}{0.04} & 0.14 & 0.10 & 0.23 \\
& \textsc{uk} & 0.07 & 0.05 & \multicolumn{1}{c|}{0.12} & 0.01 & 0.01 & \multicolumn{1}{c|}{0.02} & 0.06 & 0.05 & 0.11 \\
& \textsc{vi} & 0.21 & 0.18 & \multicolumn{1}{c|}{0.30} & 0.04 & 0.03 & \multicolumn{1}{c|}{0.06} & 0.18 & 0.15 & 0.26 \\
& \textsc{zh} & 0.00 & 0.00 & \multicolumn{1}{c|}{0.00} & 0.00 & 0.00 & \multicolumn{1}{c|}{0.00} & 0.00 & 0.00 & 0.00 \\
\hline
\multirow{13}{*}{\textsc{M2M100}}   
& \textsc{ar} & 0.20 & 0.31 & \multicolumn{1}{c|}{0.16}  & 0.08 & 0.11 & \multicolumn{1}{c|}{0.07}  & 0.19 & 0.29 & 0.15 \\
& \textsc{de} & 0.29 & 0.42 & \multicolumn{1}{c|}{0.24}  & 0.13 & 0.17 & \multicolumn{1}{c|}{0.11}  & 0.27 & 0.39 & 0.23 \\
& \textsc{es} & 0.34 & 0.46 & \multicolumn{1}{c|}{0.30}  & 0.17 & 0.22 & \multicolumn{1}{c|}{0.16}  & 0.32 & 0.43 & 0.28 \\
& \textsc{fr} & 0.28 & 0.49 & \multicolumn{1}{c|}{0.22}  & 0.13 & 0.22 & \multicolumn{1}{c|}{0.10}  & 0.26 & 0.45 & 0.21 \\
& \textsc{it} & 0.25 & 0.44 & \multicolumn{1}{c|}{0.19}  & 0.09 & 0.17 & \multicolumn{1}{c|}{0.07}  & 0.23 & 0.41 & 0.18 \\
& \textsc{ja} & 0.08 & 0.10 & \multicolumn{1}{c|}{0.07}  & 0.03 & 0.03 & \multicolumn{1}{c|}{0.03}  & 0.08 & 0.10 & 0.07 \\
& \textsc{nl} & 0.38 & 0.48 & \multicolumn{1}{c|}{0.34}  & 0.20 & 0.24 & \multicolumn{1}{c|}{0.19}  & 0.36 & 0.46 & 0.33 \\
& \textsc{pl} & 0.31 & 0.37 & \multicolumn{1}{c|}{0.29}  & 0.17 & 0.19 & \multicolumn{1}{c|}{0.16}  & 0.30 & 0.36 & 0.29 \\
& \textsc{pt} & 0.31 & 0.43 & \multicolumn{1}{c|}{0.26}  & 0.14 & 0.19 & \multicolumn{1}{c|}{0.13}  & 0.28 & 0.39 & 0.24 \\
& \textsc{sv} & 0.31 & 0.40 & \multicolumn{1}{c|}{0.28}  & 0.14 & 0.18 & \multicolumn{1}{c|}{0.14}  & 0.30 & 0.38 & 0.27 \\
& \textsc{uk} & 0.27 & 0.36 & \multicolumn{1}{c|}{0.25}  & 0.14 & 0.17 & \multicolumn{1}{c|}{0.14}  & 0.27 & 0.35 & 0.25 \\
& \textsc{vi} & 0.33 & 0.42 & \multicolumn{1}{c|}{0.30}  & 0.16 & 0.20 & \multicolumn{1}{c|}{0.15}  & 0.31 & 0.38 & 0.28 \\
& \textsc{zh} & 0.03 & 0.04 & \multicolumn{1}{c|}{0.03}  & 0.01 & 0.01 & \multicolumn{1}{c|}{0.01}  & 0.03 & 0.04 & 0.03 \\

 \hline

 \multirow{13}{*}{\textsc{mBART}}
 & \textsc{ar} & 0.15 & 0.17 & \multicolumn{1}{c|}{0.14}  & 0.10 & 0.12 & \multicolumn{1}{c|}{0.08}  & 0.16 & 0.12 & 0.14 \\
 & \textsc{de} & 0.19 & 0.23 & \multicolumn{1}{c|}{0.18}  & 0.05 & 0.06 & \multicolumn{1}{c|}{0.05}  & 0.18 & 0.22 & 0.17 \\
 & \textsc{es} & 0.32 & 0.44 & \multicolumn{1}{c|}{0.29}  & 0.16 & 0.20 & \multicolumn{1}{c|}{0.15}  & 0.30 & 0.41 & 0.27 \\
 & \textsc{fr} & 0.30 & 0.50 & \multicolumn{1}{c|}{0.24}  & 0.14 & 0.24 & \multicolumn{1}{c|}{0.12}  & 0.29 & 0.47 & 0.23 \\
 & \textsc{it} & 0.16 & 0.22 & \multicolumn{1}{c|}{0.14}  & 0.02 & 0.03 & \multicolumn{1}{c|}{0.02}  & 0.14 & 0.19 & 0.13 \\
 & \textsc{ja} & 0.04 & 0.05 & \multicolumn{1}{c|}{0.03}  & 0.00 & 0.00 & \multicolumn{1}{c|}{0.00}  & 0.04 & 0.05 & 0.03 \\
 & \textsc{nl} & 0.41 & 0.47 & \multicolumn{1}{c|}{0.39}  & 0.23 & 0.26 & \multicolumn{1}{c|}{0.23}  & 0.39 & 0.45 & 0.37 \\
 & \textsc{pl} & 0.21 & 0.23 & \multicolumn{1}{c|}{0.21}  & 0.09 & 0.09 & \multicolumn{1}{c|}{0.08}  & 0.20 & 0.22 & 0.21 \\
 & \textsc{pt} & 0.19 & 0.23 & \multicolumn{1}{c|}{0.18}  & 0.06 & 0.07 & \multicolumn{1}{c|}{0.06}  & 0.17 & 0.22 & 0.16 \\
 & \textsc{ru} & 0.26 & 0.32 & \multicolumn{1}{c|}{0.24}  & 0.11 & 0.14 & \multicolumn{1}{c|}{0.11}  & 0.26 & 0.32 & 0.23 \\
 & \textsc{sv} & 0.30 & 0.37 & \multicolumn{1}{c|}{0.27}  & 0.13 & 0.15 & \multicolumn{1}{c|}{0.12}  & 0.29 & 0.35 & 0.26 \\
 & \textsc{uk} & 0.21 & 0.27 & \multicolumn{1}{c|}{0.20}  & 0.08 & 0.11 & \multicolumn{1}{c|}{0.08}  & 0.21 & 0.26 & 0.19 \\
 & \textsc{vi} & 0.16 & 0.17 & \multicolumn{1}{c|}{0.17}  & 0.03 & 0.03 & \multicolumn{1}{c|}{0.03}  & 0.14 & 0.15 & 0.15 \\
 & \textsc{zh} & 0.01 & 0.00 & \multicolumn{1}{c|}{0.01} & 0.00 & 0.00 & \multicolumn{1}{c|}{0.00} & 0.01 & 0.03 & 0.02 \\

 \thickhline
 
\multirow{13}{*}{\textsc{mT5}} 
 & \textsc{ar} & 0.17 & 0.37 & \multicolumn{1}{c|}{0.12}  & 0.05 & 0.10 & \multicolumn{1}{c|}{0.04}  & 0.16 & 0.35 & 0.11 \\
 & \textsc{de} & 0.30 & 0.44 & \multicolumn{1}{c|}{0.25}  & 0.13 & 0.17 & \multicolumn{1}{c|}{0.12}  & 0.29 & 0.42 & 0.25 \\
 & \textsc{es} & 0.29 & 0.49 & \multicolumn{1}{c|}{0.23}  & 0.13 & 0.21 & \multicolumn{1}{c|}{0.10}  & 0.28 & 0.47 & 0.22 \\
 & \textsc{fr} & 0.28 & 0.47 & \multicolumn{1}{c|}{0.22}  & 0.11 & 0.19 & \multicolumn{1}{c|}{0.09}  & 0.26 & 0.44 & 0.20 \\
 & \textsc{it} & 0.29 & 0.49 & \multicolumn{1}{c|}{0.23}  & 0.14 & 0.22 & \multicolumn{1}{c|}{0.11}  & 0.28 & 0.46 & 0.22 \\
 & \textsc{ja} & 0.16 & 0.25 & \multicolumn{1}{c|}{0.13}  & 0.06 & 0.08 & \multicolumn{1}{c|}{0.05}  & 0.16 & 0.25 & 0.12 \\
 & \textsc{nl} & 0.32 & 0.49 & \multicolumn{1}{c|}{0.26}  & 0.13 & 0.20 & \multicolumn{1}{c|}{0.11}  & 0.30 & 0.47 & 0.25 \\
 & \textsc{pl} & 0.23 & 0.39 & \multicolumn{1}{c|}{0.18}  & 0.08 & 0.13 & \multicolumn{1}{c|}{0.07}  & 0.22 & 0.38 & 0.18 \\
 & \textsc{pt} & 0.29 & 0.42 & \multicolumn{1}{c|}{0.25}  & 0.13 & 0.17 & \multicolumn{1}{c|}{0.12}  & 0.28 & 0.40 & 0.24 \\
 & \textsc{sv} & 0.28 & 0.43 & \multicolumn{1}{c|}{0.23}  & 0.12 & 0.18 & \multicolumn{1}{c|}{0.10}  & 0.27 & 0.42 & 0.22 \\
 & \textsc{uk} & 0.22 & 0.38 & \multicolumn{1}{c|}{0.18}  & 0.08 & 0.13 & \multicolumn{1}{c|}{0.07}  & 0.22 & 0.37 & 0.17 \\
 & \textsc{vi} & 0.27 & 0.43 & \multicolumn{1}{c|}{0.23}  & 0.11 & 0.17 & \multicolumn{1}{c|}{0.10}  & 0.26 & 0.40 & 0.21 \\
 & \textsc{zh} & 0.13 & 0.22 & \multicolumn{1}{c|}{0.10}  & 0.04 & 0.06 & \multicolumn{1}{c|}{0.03}  & 0.13 & 0.21 & 0.10 \\
\thickhline
%  \hline
\end{tabular}%}
\caption{Evaluation on different models}
\label{tab:evaluation_results}
\end{table}

\begin{table}[tbh!]
    \centering

    \begin{tabular}{l|c|c}
    \thickhline
    
    Category & Num of articles &   Num of subcategories \\
    
    \thickhline
 Agriculture, food, and drink & 298 & 10 \\
Albums & 1350 & 13 \\
Architecture & 1062 & 11 \\
Art & 368 & 3 \\
Biology and medicine & 1889 & 21 \\
Chemistry and materials science & 184 & 14 \\
Classical compositions & 137 & 2 \\
Computing and engineering & 383 & 11 \\
Earth science & 1357 & 15 \\
Film & 1157 & 18 \\
Geography & 666 & 9 \\
Language and literature & 1308 & 17 \\
Mathematics and mathematicians & 110 & 3 \\
Media and drama & 657 & 6 \\
Other music articles & 878 & 7 \\
Philosophy & 216 & 6 \\
Physics and astronomy & 398 & 11 \\
Places & 533 & 10 \\
Religion & 424 & 5 \\
Royalty, nobility, and heraldry & 684 & 4 \\
Songs & 2246 & 23 \\
Television & 2586 & 113 \\
Transport & 2404 & 17 \\
World history & 1629 & 14 \\
Armies and military units & 384 & 4 \\
Baseball & 431 & 2 \\
Basketball & 251 & 2 \\
Battles, exercises, and conflicts & 1051 & 10 \\
Cricket & 139 & 2 \\
Culture, sociology, and psychology & 381 & 8 \\
Economics and business & 317 & 5 \\
Education & 280 & 3 \\
Football & 1394 & 7 \\
Hockey & 264 & 3 \\
Law & 543 & 10 \\
Magazines and print journalism & 151 & 2 \\
Military aircraft & 151 & 2 \\
Military decorations and memorials & 24 & 2 \\
Military people & 797 & 7 \\
Military ranks and positions & 7 & 1 \\
Motorsport & 317 & 2 \\
Multi-sport event & 421 & 5 \\
Other sports & 841 & 31 \\
Politics and government & 654 & 11 \\
Pro wrestling & 344 & 5 \\
Recreation & 278 & 9 \\
Video games & 1639 & 20 \\
Warships and naval units & 1761 & 3 \\
Weapons, equipment, and buildings & 336 & 4 \\

    \thickhline
    \end{tabular}     

  \caption{English Good articles divided into categories}
  \label{tab:wiki_stat}
\end{table}

\section{Conclusion}
\label{sec:conclusion}
We proposed a novel dataset for cross-lingual summarization. It is comparable in size to the existing largest one, while being more broad in topics and including longer documents and summaries. We have evaluated several well known models for abstractive summarization on this dataset and found out that the performance is stronger correlated with the language itself than the model. E.g. the Dutch language has better scores on average for all the models. We hypothesize that this reflects the culture of Wikipedia writing in Dutch language, rather than the language structure.

We hope that this dataset will ease the way for other researchers in the field of cross-lingual summarization.

\end{document}